\documentclass[a4paper,11pt]{article}
\usepackage{graphicx}
\usepackage{amsfonts}
\usepackage{amsmath}
\usepackage[utf8]{inputenc}
\usepackage{url}
\usepackage{color}
 \usepackage{amssymb}
 \usepackage{amsthm}
\usepackage{multirow}
\usepackage{overpic}

\usepackage{bbm, dsfont}

\newtheorem {Proposition}{Proposition}[section]

\newtheorem {theo}[Proposition]{Theorem}

\numberwithin{equation}{section}

\newcommand{\p}{\mathbb{P}}
\newcommand{\R}{\mathbb{R}}

\newcommand\BibTeX{{\rmfamily B\kern-.05em \textsc{i\kern-.025em b}\kern-.08em
T\kern-.1667em\lower.7ex\hbox{E}\kern-.125emX}}

\title{Confidence Intervals for Testing Disparate Impact in Fair Learning}

 \author{Philippe Besse$^b$, Eustasio del Barrio$^a$, Paula Gordaliza$^{a,b}$, and Jean-Michel Loubes$^b$\\
	$^{a}$\textit{IMUVA, Universidad de Valladolid}\\
	$^{b}$\textit{Institut de Math\'ematiques de Toulouse
}}

\begin{document}

\maketitle

\begin{abstract}
We provide the asymptotic distribution of the major indexes used in the statistical literature to quantify disparate treatment in machine learning. We aim at promoting the use of confidence intervals when testing the so-called group disparate impact. We illustrate on some examples the importance of using confidence intervals and not a single value.
\end{abstract}

{\bf  keywords : } {Test, Fair Learning}

\section{Introduction}
With the generalization of machine learning algorithms in a large variety of fields, their impact on human life is growing over the years. Originally designed to improve recommendation systems in the internet industry, they are now widely used in a large number of  very sensitive areas such as medicine, human ressources, banks and insurance, criminal justice risk assessment, see for instance in \cite{romei_ruggieri_2014}, \cite{berk2017fairness} \cite{pedreschi2012study} or \cite{2018arXiv180204422F} and references therein. Meant to take automatic accurate and efficient decisions mimicking and even outmatching human expertise,  machine learning algorithms may exhibit discriminatory behaviours in the sense that groups of populations are treated in  distinct ways. Even if some discrimination may appear naturally and could be thought as acceptable (see for instance in \cite{5694053}), quantifying the effect of machine learning with respect to a given situation is of high importance.  This notion of fairness in machine learning algorithms has received a growing interest over the last years and is crucial in order to  guarantee a fair treatment to all population. Moreover, enhancing fairness can also contribute to a better trust for machine learning algorithms in the whole population. \vskip .1in
Yet providing a definition of fairness  or equity in machine learning algorithms is a complicated task and several propositions have been formulated. We will focus on the issue of biased training data, which is one of the several possible causes of discriminatory outcomes in
machine learning. First described in terms of law \cite{winrow2009disparity}, fairness is now quantified in order to detect lack of fairness in automatic algorithms. According to the objectives, quantitative measures of fairness have been designed but  these measures rest on
unstated assumptions about fairness in society.  Fairness is often defined with respect to selected attributes called protected attribute which represents a discriminatory information in the population that should not be used or retrieved in the algorithm decision.  Among all these criteria, two main categories have been considered. The first one deals with the repartition of a decision rule with respect to the protected attribute. This point of view gives rise to the Disparate Impact described for instance in \cite{FFMSV}. The second one tackles the issue of disparate error rates of the algorithmic decisions between the different groups of the population. This point of view has been originally proposed for recidivism of defendants in \cite{flores2016false}. Many others criteria (see for instance in \cite{berk2017fairness} for a review) have been proposed leading sometimes to incompatible formulations as stated in \cite{2017arXiv170300056C}. Note finally that the notion of fairness is closely related to the notion of privacy as pointed out in \cite{2011arXiv1104.3913D}.  \vskip .1in
Our goal in this paper is not to discuss the criterions chosen but rather to promote the use of confidence intervals to control the risk of false discriminatory assessment. If many criteria have been described in the literature  of fair learning, they are often used as a score without statistical control. In the cases where test procedures or  confidence bounds are provided, they are obtained  using a resampling scheme to get standardized Gaussian confidence intervals under a Gaussian assumption which does not correspond to the distribution of the observations. Hence in this work, we provide the exact asymptotic distribution of the estimates of some  fairness criteria obtained through the classical approach of the Delta method described in \cite{van1998asymptotic}.

\section{Quantifying unfair treatment using Statistical Criteria}
Even if unfairness in machine learning is a recent topic, many criteria have been already considered in order detect unfair algorithmic treatment.\\
 In the literature, detecting unfair treatment can be first be achieved by looking at individual outcomes and measuring how different they might be for similar persons. For this, one may be interested by quantifying how dissimilar outcomes may be encountered in a neighbourhood of a person that might have suffered a biased decision.  This is usually achieved by looking at the prediction of the algorithm for two individuals in every characteristic except the value of a variable which may lead to possible disparate treatment. Yet such measures may be very unstable and not representative of the whole behaviour of the decision rule.\\
 \indent For these reasons, statistical measures that detect group discriminations or group disparate treatment by an algorithm have been recently introduced to assess unfair algorithmic treatment with respect to a variable $S$ called protected variable. \\
 Actually, the statistical model is the following. The problem  consists in  forecasting a binary variable $Y\in \left\lbrace0,1\right\rbrace$, using observed covariates  $X \in \R^d, \ d\geqslant 1.$  We assume moreover that the population can be divided into two categories that represent a bias,  modeled by a  variable $S \in \{0,1\}$. This variable is called the protected attribute, which takes the values $S=0$ for the \textquotedblleft minority\textquotedblright class and supposed to be the unfavored class, and $S=1$ for the \textquotedblleft default\textquotedblright, and usually favored class. $S$ represents the group we wish to protect from discrimination, and the bias represents the degree to which they have been discriminated against. Note that in the case where $S$ is not a binary variable but multidimensional and multi-class, we can perform several tests identifying in each case a less favored case.  We  introduce also a notion of positive prediction in the sense that $Y=1$ represents a success while $Y=0$ is a failure.  Hence the classification problem aims at predicting a success from the variables $X$, using a family of binary classifiers $g\in \mathcal{G}: \R^d \rightarrow \{0,1\}$. For every $ g\in\mathcal{G}$, the outcome of the classification will be the prediction $\hat{Y}=g(X)$.

The different  frameworks  considered in the statistical literature  intend to quantify the distance  between the outcome of the algorithm and an ideal situation where decisions should not be impacted by the protected variable. All criteria amount to measure how the decision is correlated to the clustering of the population obtained using the protected variable.  Yet they differ  depending on the observations which are available to the statistician, which gives rise to the following criteria. 

\begin{itemize}
	\item When considering a database made of observations $X$ and a variable $Y \in \{0,1\}$ to be predicted by a classifier $g$, the disparate impact measures how the two labels are spread between the subgroups defined by the variable $S$. Namely, knowing the decision $g(X)$ and the protected variable $Y$, the disparate impact assessment DIA of this classifier is defined as
$$DIA(g)=\frac{\mathbb{P}(g(X)=1 | S=0)}{ \mathbb{P}(g(X)=1 | S=1)}. $$

This quantity quantifies how far a classifier is from the ideal situation, called the Statistical parity where
$$\mathbb{P}(g(X)=1 | S=0) = \mathbb{P}(g(X)=1 | S=1). $$ This means that the probability of a successful outcome is the same across the groups. For instance, if we consider that the protected variable represents gender, the value $S=0$ would be assigned to \textquotedblleft female\textquotedblright and $S=1$ to \textquotedblleft male\textquotedblright, we would say that the algorithm used by a company achieves \textit{Statistical Parity} if a man and a woman have the same probability of success (for instance being hired or promoted). The classifier $g: \R^d \rightarrow \{0,1\}$ is said to have Disparate Impact at level $\tau \in \left(0,1\right]$, with respect to $(X,S)$,
 if $DI(g,X,S) \leq \tau$. Note the Disparate Impact of a classifier measures its level of fairness: the smaller the value of $\tau$, the less fair it is. The classification rules considered in this framework are such that $b(g) \geqslant a(g)>0,$ because we are assuming that the default class $S=1$ is more likely to have a successful outcome. Thus, in the definition, the level of fairness $\tau$ takes values $0 < \tau \leq 1$. We point out that the value $\tau_0=0.8=4/5$, which is also known in the literature as the \textit{$80 \%$ rule} has been cited as a legal score to decide whether the discrimination of the algorithm is acceptable or not (see for instance in \cite{FFMSV}). This rule can be explained as \textquotedblleft for every 5 individuals with successful outcome in the majority class, 4 in the minority class will have a successful outcome too\textquotedblright.\\\indent  In what follows, to promote fairness, it will be useful to state the definition in the reverse sense. A classifier does not have Disparate Impact at level $\tau$, with respect to $(X,S)$, if $DI(g,X,S)>\tau$. 

Within this framework, the disparate impact assessment can be compared to the Disparate Impact
$$DI(Y):=\frac{\mathbb{P}(Y=1 | S=0)}{\mathbb{P}(Y=1 | S=1)},$$ which quantifies the same bias but on the true distribution of the label $Y$. It can be useful to determine whether a decision rule increases or not the discrimination that exists in the learning sample.
 \item When $Y$, $g(X)$ and $S$ can be observed, other criterion can be used to give an insight on possible unfairness.

For this, fairness can be defined as the situation where the accuracy of the classification process is the same for both groups. We point out that this is equivalent to assess that the false negative rate and the false positive rate are the same for both groups. The mathematical formalization of this statement is given by

$$ \begin{cases} \mathbb{P}(g(X)=1 |Y=1, S=0) & = \mathbb{P}(g(X)=1 |Y=1, S=1) \\

\mathbb{P}(g(X)=0 |Y=0, S=0) & = \mathbb{P}(g(X)=0 |Y=0, S=1) \end{cases}$$
This situation is called Conditional Procedure Accuracy Equality. We point out that this is equivalent to
 quantify the amount of fairness. For this  we consider for a given classifier $g$ the quantity that false negative rate and false positive rate are the same for both groups, which is also called equalized odds.  This framework was developed in the COMPAS controversy, see for instance in \cite{flores2016false} or \cite{compas} to forecast recidivism of prisoners.  When an offender is eligible for parole, judges assess the likelihood that the offender will re-offend after being released as part of the parole decision. Many jurisdictions now use automated prediction methods like the COMPAS score which are taylored to get a balanced Disparate Impact. Yet the disparate treatment may arise from a disparity in the errors of the decision rule.  \\
To quantify the difference with respect to this fair situation we will consider the following quantities
$$CA_1(g)= \frac{\mathbb{P}(g(X)=1 |Y=1, S=0)}{\mathbb{P}(Y=1, S=0)} /  \frac{\mathbb{P}(g(X)=1 |Y=1, S=1)}{\mathbb{P}(Y=1, S=1)}.$$                                                            
$$CA_0(g)= \frac{\mathbb{P}(g(X)=0 |Y=0, S=0)}{\mathbb{P}(Y=0, S=0)} /  \frac{\mathbb{P}(g(X)=0 |Y=0, S=1)}{\mathbb{P}(Y=0, S=1)}.$$        
\item In the same setting where $Y,g(X),S$ are observed, another criterion of fairness is given by the Conditional Use Accuracy Equality. This amounts to define fairness as the situation where the use, conditionally to the algorithm outcome, is the same for both groups. 
$$ \begin{cases}\mathbb{P}(Y=1 | g(X)=1, S=0) & = \mathbb{P}(Y=1 | g(X)=1, S=1)\\
\mathbb{P}(Y=0 | g(X)=0, S=0) & = \mathbb{P}(Y=0 | g(X)=0, S=1) .\end{cases} $$
  This criterion corresponds to the case where the difference between the use of the classifier and the real label is measured between both groups. Here again we will use the following criteria to assess the  gap between the data and the conditional use accuracy equality 
  $$CU_1(g)= \frac{\mathbb{P}(Y=1 |g(X)=1, S=0)}{\mathbb{P}(g(X)=1, S=0)} /  \frac{\mathbb{P}(Y=1 |g(X)=1, S=1)}{\mathbb{P}(g(X)=1, S=1)}.$$                                                            
$$CU_0(g)= \frac{\mathbb{P}(Y=0 |g(X)=0, S=0)}{\mathbb{P}(g(X)=0, S=0)} /  \frac{\mathbb{P}(Y=0 |g(X)=0, S=1)}{\mathbb{P}(g(X)=0, S=1)}.$$
This criteria may be seen close to the previous case but it focuses on the odds of each individual while previous is more devoted to the analysis of the classification algorithm.                                                    
\end{itemize}
Such criteria are imperfect and may fail to capture all  aspects of fairness. In particular, it is easy to achieve statistical parity simply by flipping the
labels of an arbitrary set of individuals in the protected class or randomly repairing the data by pushing a random part of the data towards the so-called Wasserstein barycenter of the data as described in \cite{MyFairPaper}. Yet they provide some quantification of a level of unfair treatment and give some insights about the disparate treatment received by the different groups 
\section{Testing lack of fairness and confidence intervals}
Let $\left(X_i, S_i, \hat{Y}_i=g(X_i)\right), i=1, \ldots, n,$ be a random sample of independent and equally distributed variables. Previous criterion can be consistently estimated by their empirical version. Yet the value of the criterion may depend on the data sample. Due to the importance of obtaining an accurate proof of  unfairness in a decision rule it is important to obtain confidence intervals in order to control the error of detecting unfairness. In the  literature it is often achieved by computing the mean over several sampling of the data.   We provide in the following the exact asymptotic behaviors of the estimates in order to build confidence intervals. \vskip .1in

\begin{theo}[Asymptotic behavior of the Disparate Impact Assessment estimator]
 \label{th:DI}
Set the empirical estimator of DI(g) as
\begin{equation*}
T_n := \displaystyle\frac{\sum_{i=1}^{n}\mathbbm{1}_{g(X_i)=1}\mathbbm{1}_{S_i=0}\sum_{i=1}^{n}\mathbbm{1}_{S_i=1}}{\sum_{i=1}^{n}\mathbbm{1}_{g(X_i)=1}\mathbbm{1}_{S_i=1}\sum_{i=1}^{n}\mathbbm{1}_{S_i=0}}.
\end{equation*} Then the asymptotic distribution of this quantity is given by 
\begin{equation}\label{eq:convergenceTn}
\frac{\sqrt{n}}{\sigma}\left(T_n-DI(g,X,S)\right) \xrightarrow{d} N(0,1),  \ as \ n\rightarrow \infty,
\end{equation}
where $\sigma = \displaystyle\sqrt{\nabla\varphi^T\left(\mathbb{E}Z_1\right) \Sigma_4 \nabla\varphi\left(\mathbb{E}Z_1\right)}$ and 
\begin{equation*}
\nabla\varphi^T\left(\mathbb{E}Z_1\right)=\left(\frac{\pi_1}{p_1\pi_0},-\frac{p_0\pi_1}{p_1^2\pi_0}, -\frac{p_0\pi_1}{p_1\pi_0^2},\frac{p_0}{p_1\pi_0} \right)
\end{equation*}
\begin{equation*}
\Sigma_4= \left(\begin{array}{cccc}
p_0(1-p_0) & & & \\
-p_0p_1 & p_1(1-p_1) & & \\
\pi_1p_0 & -\pi_0p_1 & \pi_0\pi_1 & \\
-\pi_1p_0 & \pi_0p_1 & -\pi_0\pi_1 & \pi_0\pi_1
\end{array} \right),
\end{equation*}
where we have denoted $\pi_s=\p(S_1=s)$ and $p_s=\p(g(X_1)=1, S_i=s), \ s=0,1,$ .
\end{theo}
Hence, we can provide a confidence interval when estimating the disparate impact over a data set. Actually $\left(T_n \pm \frac{\sigma}{\sqrt{n}}Z_{1-\frac{\alpha}{2}}\right)$ is a confidence interval for the parameter $DI(g,X,S)$ asymptotically of level $1-\alpha$. \\ Previous theorem can be used to test the presence of disparate impact at a given level.
\begin{equation}
H_{0, \beta} : \ DI(g,X,S) \leqslant \beta \ \ \ vs. \ \ \ H_{1, \beta} : \ DI(g,X,S) > \beta
\end{equation}
aims at checking if $g$ has Disparate Impact at level $\beta$.
We want to check wether $ DI(g,X,S) \leq  \beta$.
Under $H_0$, the inequality $T_n-\beta \leqslant T_n -DI(g,X,S)$ holds, and so
\begin{equation*}
\frac{\sqrt{n}}{\sigma}\left(T_n-\beta\right) \leqslant \frac{\sqrt{n}}{\sigma} \left(T_n -DI(g,X,S)\right).
\end{equation*}
Finally, from the inequality above and (\ref{eq:convergenceTn}), we have that
\begin{equation*}\label{equ:levelacceptanceH0}
\p_{H_0}\left(\frac{\sqrt{n}}{\sigma}\left(T_n- \beta \right) < Z_{1-\alpha} \right) \geqslant \p_{H_0}\left(\frac{\sqrt{n}}{\sigma}\left(T_n- DI(g,X,S) \right) < Z_{1-\alpha} \right) \longrightarrow 1- \alpha,  \ as \ n\rightarrow \infty,
\end{equation*}
and, equivalently,
\begin{equation*}
\p_{H_0}\left(\frac{\sqrt{n}}{\sigma}\left(T_n-\beta\right) \geqslant Z_{1-\alpha} \right) \leqslant \p_{H_0}\left(\frac{\sqrt{n}}{\sigma}\left(T_n- DI(g,X,S) \right) \geqslant Z_{1-\alpha} \right) \longrightarrow \alpha,  \ as \ n\rightarrow \infty,
\end{equation*}
where $Z_{1-\alpha}$ is the $(1-\alpha)$-quantile of $N(0,1)$. In conclusion, the test rejects $H_0$ at level $\alpha$ when
\begin{equation*}
\p_{H_0}\left(\frac{\sqrt{n}}{\sigma}\left(T_n-\beta\right) \geqslant Z_{1-\alpha} \right) \geqslant \alpha.
\end{equation*}

The proof  of this theorem is quite classical and is postponed to the Appendix.\vskip .1in
When dealing with Conditional Accuracy, we want to study the asymptotic behavior of the estimators of the rates of the True Positives and True Negatives across both groups. The reasoning is similar for the two quantities $CA_1(g)$ and $CA_0(g)$, so we will only show the convergence of the True Positive Assessment estimator.

\begin{theo} \label{th:CE}
Set an estimate of $CA_1(g)$
\begin{equation*}
R_n := \displaystyle\frac{\sum_{i=1}^{n}\mathbbm{1}_{g(X_i)=1}\mathbbm{1}_{Y_i=1}\mathbbm{1}_{S_i=0}\sum_{i=1}^{n}\mathbbm{1}_{Y_i=1}\mathbbm{1}_{S_i=1}}{\sum_{i=1}^{n}\mathbbm{1}_{g(X_i)=1}\mathbbm{1}_{g(X_i)=1}\mathbbm{1}_{S_i=1}\sum_{i=1}^{n}\mathbbm{1}_{Y_i=1}\mathbbm{1}_{S_i=0}}.
\end{equation*}
Then, the assymptotic distribution of this quantity is given by
\begin{equation}\label{equ:convergenceTnCA1}
\frac{\sqrt{n}}{\sigma}\left(R_n-CA_1(g)\right) \xrightarrow{d} N(0,1),  \ as \ n\rightarrow \infty,
\end{equation}
where $\sigma = \displaystyle\sqrt{\nabla\varphi^T\left(\mathbb{E}Z_1\right) \Sigma_4 \nabla\varphi\left(\mathbb{E}Z_1\right)}$
and
\begin{equation*}
\nabla\varphi^T\left(\mathbb{E}Z_1\right)=\left(\frac{r_1}{p_1r_0},-\frac{p_0r_1}{p_1^2r_0}, -\frac{p_0r_1}{p_1r_0^2},\frac{p_0}{p_1r_0} \right)
\end{equation*}
\begin{equation*}
\Sigma_4= \left(\begin{array}{cccc}
p_0(1-p_0) & & & \\
-p_0r_1 & p_1(1-p_1) & & \\
p_0(1-r_0) & -p_1r_0 & r_0(1-r_0) & \\
p_0r_1 & p_1(1-r_1) & -r_0r_1 & r_1(1-r_1)
\end{array} \right),
\end{equation*}
where we have denoted $p_s=\p(g(X_1)=1, Y_1=1, S_1=s),$ and $r_s=\p(Y_1=1,S_1=s), \ s=0,1.$
\end{theo}

 Again, we will give the theorem that establish the asymptotic behaviour of the $CU_1(g)$ estimator, noting that the corresponding to the other $CU_0(g)$ is analogously analyzed.
\begin{theo} \label{th:CUAE}
	Set an estimate of $CU_1(g)$
	\begin{equation*}
	U_n := \displaystyle\frac{\sum_{i=1}^{n}\mathbbm{1}_{g(X_i)=1}\mathbbm{1}_{Y_i=1}\mathbbm{1}_{S_i=0}\sum_{i=1}^{n}\mathbbm{1}_{g(X_i)=1}\mathbbm{1}_{S_i=1}}{\sum_{i=1}^{n}\mathbbm{1}_{g(X_i)=1}\mathbbm{1}_{Y_i=1}\mathbbm{1}_{S_i=1}\sum_{i=1}^{n}\mathbbm{1}_{g(X_i)=1}\mathbbm{1}_{S_i=0}}.
	\end{equation*}
	Then, the assymptotic distribution of this quantity is given by
	\begin{equation}\label{equ:convergenceTnCU1}
	\frac{\sqrt{n}}{\sigma}\left(U_n-CU_1(g)\right) \xrightarrow{d} N(0,1),  \ as \ n\rightarrow \infty,
	\end{equation}
	where $\sigma = \displaystyle\sqrt{\nabla\varphi^T\left(\mathbb{E}Z_1\right) \Sigma_4 \nabla\varphi\left(\mathbb{E}Z_1\right)}$
	and
	\begin{equation*}
	\nabla\varphi^T\left(\mathbb{E}Z_1\right)=\left(\frac{r_1}{p_1r_0},-\frac{p_0r_1}{p_1^2r_0}, -\frac{p_0r_1}{p_1r_0^2},\frac{p_0}{p_1r_0} \right)
	\end{equation*}
	\begin{equation*}
	\Sigma_4= \left(\begin{array}{cccc}
	p_0(1-p_0) & & & \\
	-p_0r_1 & p_1(1-p_1) & & \\
	p_0(1-r_0) & -p_1r_0 & r_0(1-r_0) & \\
	p_0r_1 & p_1(1-r_1) & -r_0r_1 & r_1(1-r_1)
	\end{array} \right),
	\end{equation*}
	where we have denoted $p_s=\p(g(X_1)=1, Y_1=1, S_1=s),$ and $r_s=\p(g(X_1)=1,S_1=s), \ s=0,1.$
\end{theo}
The proof of this theorem is similar to the one of Theorem~\ref{th:CE} and is omitted.

\section{Using confidence Intervals for real dataset} \label{s:use}
To illustrate these tests we first consider the \textit{Adult Income} data set. It contains $29.825$ instances consisting in the values of $14$ attributes, $6$ numeric and $8$ categorical, and a categorization of each person as having an income of more or less than $50,000\$$ per year. This attribute will be the target variable in the study.  We have access to the whole information, the variables $X$, the true observed variable $Y$. Two variables can be considered as protected : the sex and the origin. \vskip .1in
We first estimate the Disparate Impact that describes the discrimination in the learning sample with respect to both variables. This score will describe how the learning sample presents a group discrimination either due to the selection of the sample or the discrimination present in the whole population. Define $ {DI}_S$ (respectively ${DI}_O$)the Disparate Impact with respect to the sex variable such that $S=0$ corresponds to female while $S=1$ corresponds to male
(respectively with respect to the origin variables $O=0$ corresponds to foreign origin while $O=1$ corresponds to native). We get  the following values with their corresponding confidence interval at level $5\%$.
$$ {DI}_S = 0.36  \in \quad   [0.34 ,0.39 ],$$
$$ {DI}_O=0.6  \in  \quad [0.55 , 0.64 ].$$
If we consider the threshold described in the  proofs of discriminate behaviors in the former USA trials (see for instance in~\cite{mercat2016discrimination} and references therein), the discriminate impact should be greater than 0.8 to guarantee no discriminate impact. Hence in the situation, both variables generate discrimination, in a more severe way for the sex than for the origin. \\
We now consider learning algorithms to predict the variable of interest and study the disparate impact of these decision rules.  For this we consider  either a logit model or a random forest built with all variables including the protected variables and optimized using cross validation methods.  We obtain the following results, 
$$ DI_S({\rm logit}) =  0.25  \in \quad [    0.21 , 0.3 ];  \quad DI_S({\rm RF})= 0.32 \in \quad [     0.27,  0.37 ]$$
$$ DI_O({\rm logit}) = 0.51 \in \quad [    0.41 , 0.6 ]; \quad DI_O({\rm RF})= 0.58  \in \quad [     0.52,  0.61 ]$$
We can see that in both cases, the algorithms enforce the discrimination by having smaller  disparate impact than for the true variable. Actually classification algorithms aimed at discriminating  the population, enhancing the bias found in the sample. Hence previous algorithms are unfair in the sense that discrimination is reinforced. \\
 Then, we process the same calculations with algorithms built without using the protected variables, which could correspond to a naive answer to promote fairness.
$$  DI_S({\rm logit}) =  0.27  \in \quad [    0.23 , 0.31 ];  \quad DI_S({\rm RF})= 0.32 \in \quad [     0.28,  0.38 ]    $$
$$ DI_O({\rm logit}) = 0.56 \in \quad [    0.46 , 0.66 ]; \quad DI_O({\rm RF})= 0.58  \in \quad [     0.48,  0.68 ]$$
Even if the disparate impact is improved very slightly, we observe that the changes in the disparate impact and their confidence intervals are not statistically significant.  So discarding the protected variables when building the model does not improve fairness of the predictor. Hence social determinism is stronger than the protected arguments. A woman or a non caucasian person is expected to earn less whatever its education level. This justifies the use of fairness  mathematical methods to reduce disparate treatment as discussed in  \cite{kleinberg2016inherent} or \cite{MyFairPaper} for instance.  \vskip .1in
The second data set is German Credit Data Set. This dataset is often claimed to exhibit some origin discrimination in the success of being given a credit  by the German bank. Hence we compute the disparate impact w.r.t Origin. We obtain
$$ DI = 0.77 \in  \quad [0.68 , 0.87 ] .$$
Hence here confidence intervals play an important role. Actually the disparate impact is not statistically significantly lower than 0.8, which entails that the discrimination of the decision rule of the German bank can not be shown, which promotes the use of a proper confidence interval.   \vskip .1in
A third  data set is composed by the data of the controversial COMPAS score detailed in \cite{dieterich2016compas}.  The data is composed of 7214 offenders with personal variables observed over two years. A score predicts their level of dangerosity which determines whether they can be released while a variable points out if there has been recidivism. Hence Recidivism of offenders is predicted using a score and confronted to possible racial discrimination which corresponds to the protected attribute. The protected variable separates the population into caucasian and non caucasian.  To evaluate the level of discrimination we first compute the disparate impact with respect to the true variable and the COMPAS score seen as a predictor. 
$$ DI= 0.76 \in [.72,.81]; \quad DI({\rm COMPAS})= 0.71 \in [0.68;0.74].$$ 
In both cases, the data are biased but the level of discrimination is low. Yet as mentioned in al the studies on this data set, the level of errors of prediction is significantly different according to the ethnic origin of the defender. Actually the conditional accuracy scores and their corresponding confidence intervals show clearly the unbalance treatment received by both populations.
$$
TPR= 0.6 \in [ 0.54,	 	0.65 ]$$
$$TNR=3.38 \in [2.46,4.3 ]$$
This unbalanced treatment is clearly assesed with the confidence interval.
\section{Conclusions}
Quantifying the level of fairness of a learning sample or of an algorithm is a difficult task since the points of views may differ to define the notion of disparate treatment. Yet, when dealing with the main indexes that has been used, it is important as in any statistical analysis to obtain a confidence interval at given level and not a single numerical value. For this we provided the asymptotic distribution of the estimates of three major fairness indexes in order to promote their use in assessing unfair treatment in machine learning algorithms. 
\section{Appendix}
Proof of Theorem~\ref{th:DI}
\begin{proof} 
Consider for $i=1,\ldots,n,$ the random vectors
\begin{equation*}
Z_i=\left( \begin{array}{c}
\mathbbm{1}_{g(X_i)=1}\mathbbm{1}_{S_i=0}\\
\mathbbm{1}_{g(X_i)=1}\mathbbm{1}_{S_i=1}\\
\mathbbm{1}_{S_i=0}\\
\mathbbm{1}_{S_i=1}
\end{array} \right),
\end{equation*}
where $\mathbbm{1}_{g(X_i)=1}\mathbbm{1}_{S_i=s} \sim B(\p(g(X_i)=1,S_i=s))$ and $\mathbbm{1}_{S_i=s} \sim B(\p(S_i=s)), \ s=0,1,$. Thus, $Z_i$ has expectation
\begin{equation*}
\mathbb{E}Z_i=\left( \begin{array}{c}
\p(g(X_i)=1, S_i=0)\\
\p(g(X_i)=1, S_i=1)\\
\p(S_i=0)\\
\p(S_i=1)
\end{array} \right).
\end{equation*}
The elements of the covariance matrix $\Sigma_4$ of $Z_i$ are computed as follows:
\begin{align*}
Cov\left(\mathbbm{1}_{g(X_i)=1}\mathbbm{1}_{S_i=0}, \mathbbm{1}_{g(X_i)=1}\mathbbm{1}_{S_i=1}\right)= \mathbb{E}\left(\mathbbm{1}^2_{g(X_i)=1}\mathbbm{1}_{S_i=0}\mathbbm{1}_{S_i=1}\right)-\p(g(X_i)=1,S_i=0)\p(g(X_i)=1,S_i=1)
\end{align*}
\begin{align*}
Cov\left(\mathbbm{1}_{g(X_i)=1}\mathbbm{1}_{S_i=0}, \mathbbm{1}_{S_i=0}\right)&= \mathbb{E}\left(\mathbbm{1}_{g(X_i)=1}\mathbbm{1}^2_{S_i=0}\right)-\p(g(X_i)=1,S_i=0)\p(S_i=0)\\
&=\p(g(X_i)=1)\p(S_i=0)- \p(g(X_i)=1,S_i=0)\p(S_i=0)\\
&=\left[1- \p(S_i=0)\right]\p(g(X_i)=1,S_i=0)
\end{align*}
\begin{multline*}
Cov\left(\mathbbm{1}_{g(X_i)=1}\mathbbm{1}_{S_i=0}, \mathbbm{1}_{S_i=1}\right)=\mathbb{E}\left(\mathbbm{1}_{g(X_i)=1}\mathbbm{1}_{S_i=0}\mathbbm{1}_{S_i=1}\right)-\p(g(X_i)=1,S_i=0)\p(S_i=1)
\end{multline*}
\begin{multline*}
Cov\left(\mathbbm{1}_{g(X_i)=1}\mathbbm{1}_{S_i=1}, \mathbbm{1}_{S_i=0}\right)=\mathbb{E}\left(\mathbbm{1}_{g(X_i)=1}\mathbbm{1}_{S_i=0}\mathbbm{1}_{S_i=1}\right)-\p(g(X_i)=1,S_i=1)\p(S_i=0)
\end{multline*}
\begin{align*}
Cov\left(\mathbbm{1}_{g(X_i)=1}\mathbbm{1}_{S_i=1}, \mathbbm{1}_{S_i=1}\right)&= \mathbb{E}\left(\mathbbm{1}_{g(X_i)=1}\mathbbm{1}^2_{S_i=1}\right)-\p(S_i=1)\p(g(X_i)=1,S_i=1)\\
&=\p(g(X_i)=1,S_i=1)- \p(S_i=1)\p(g(X_i)=1,S_i=1)\\
&=\p(g(X_i)=1,S_i=1)\left[1- \p(S_i=1)\right]\\
&=\p(g(X_i)=1,S_i=1)\p(S_i=0)
\end{align*}
and finally,
\begin{equation*}
Cov(\mathbbm{1}_{S_i=0},\mathbbm{1}_{S_i=1})=\mathbb{E}\left(\mathbbm{1}_{S_i=0}\mathbbm{1}_{S_i=1}\right)-\p(S_i=0)\p(S_i=1)=-\p(S_i=0)\p(S_i=1).
\end{equation*}
From the Central Limit Theorem in dimension 4, we have that
\begin{equation*}
\sqrt{n}\left(\bar{Z}_n-\mathbb{E}Z_1\right) \xrightarrow{d} N_4\left(\mathbf{0}, \Sigma_4\right), \ as \ n\rightarrow \infty.
\end{equation*}
Now consider the function
\begin{center}
	$
	\begin{array}{cccc}
	\varphi:& \R^4 & \longrightarrow  &\R\\
	&(x_1,x_2,x_3,x_4) & \longmapsto & \displaystyle\frac{x_1x_4}{x_2x_3}
	\end{array}
	$
\end{center}
Applying the Delta-Method (see in~\cite{van1998asymptotic}) for the function $\varphi$, we conclude that
\begin{equation*}
\sqrt{n}\left(\varphi(\bar{Z}_n)-\varphi(\mathbb{E}Z_1)\right) \xrightarrow{d} \nabla\varphi^T\left(\mathbb{E}Z_1\right)N_4\left(\mathbf{0}, \Sigma_4\right), \ as \ n\rightarrow \infty,
\end{equation*}
where $\varphi(\bar{Z}_n)= T_n, \ \varphi(\mathbb{E}Z_1)=DI(g,X,S)$. \end{proof}

Proof of Theorem~\ref{th:CE}
\begin{proof} The proof follows the same guidelines of previous proof. We set here
	 \begin{equation*}
	 Z_i=\left( \begin{array}{c}
	 \mathbbm{1}_{g(X_i)=1}\mathbbm{1}_{Y_i=1}\mathbbm{1}_{S_i=0}\\
	 \mathbbm{1}_{g(X_i)=1}\mathbbm{1}_{Y_i=1}\mathbbm{1}_{S_i=1}\\
	 \mathbbm{1}_{Y_i=1}\mathbbm{1}_{S_i=0}\\
	 \mathbbm{1}_{Y_i=1}\mathbbm{1}_{S_i=1}
	 \end{array} \right),
	 \end{equation*}
	 where $\mathbbm{1}_{g(X_i)=1}\mathbbm{1}_{Y_i=1}\mathbbm{1}_{S_i=s} \sim B(\p(g(X_i)=1,Y_i=1,S_i=s))$ and $\mathbbm{1}_{Y_i=1}\mathbbm{1}_{S_i=s} \sim B(\p(Y_i=1,S_i=s)), \ s=0,1,$. 
	 From the Central Limit Theorem, we have that
	 \begin{equation*}
	 \sqrt{n}\left(\bar{Z}_n-\mathbb{E}Z_1\right) \xrightarrow{d} N_4\left(\mathbf{0}, \Sigma_4\right), \ as \ n\rightarrow \infty.
	 \end{equation*}
	 with 
	 	 \begin{equation}
	 \Sigma_4= \left(\begin{array}{cccc}
	 p_0(1-p_0) & & & \\
	 -p_0r_1 & p_1(1-p_1) & & \\
	 p_0(1-r_0) & -p_1r_0 & r_0(1-r_0) & \\
	 p_0r_1 & p_1(1-r_1) & -r_0r_1 & r_1(1-r_1)
	 \end{array} \right).
	 \end{equation}
	 Now consider the function
	 \begin{center}
	 	$
	 	\begin{array}{cccc}
	 	\varphi:& \R^4 & \longrightarrow  &\R\\
	 	&(x_1,x_2,x_3,x_4) & \longmapsto & \displaystyle\frac{x_1x_4}{x_2x_3}
	 	\end{array}
	 	$
	 \end{center}
	 Applying the Delta-Method for the function $\varphi$, we conclude that
	 \begin{equation*}
	 \sqrt{n}\left(\varphi(\bar{Z}_n)-\varphi(\mathbb{E}Z_1)\right) \xrightarrow{d} \nabla\varphi^T\left(\mathbb{E}Z_1\right)N_4\left(\mathbf{0}, \Sigma_4\right), \ as \ n\rightarrow \infty,
	 \end{equation*}
	 where $\varphi(\bar{Z}_n)= R_n,$ and $\varphi(\mathbb{E}Z_1)=CA_1(g)$.
\end{proof}

\end{document}